\documentclass{article}

\usepackage{arxiv}

\usepackage[utf8]{inputenc} % allow utf-8 input
\usepackage[T1]{fontenc}    % use 8-bit T1 fonts
\usepackage{hyperref}       % hyperlinks
\usepackage{url}            % simple URL typesetting
\usepackage{booktabs}       % professional-quality tables
\usepackage{amsfonts}       % blackboard math symbols
\usepackage{nicefrac}       % compact symbols for 1/2, etc.
\usepackage{microtype}      % microtypography
\usepackage{lipsum}         % Can be removed after putting your text content
\usepackage{graphicx}
\usepackage[numbers,sort&compress]{natbib}
\usepackage{doi}

\usepackage{algorithmic}
\usepackage{graphicx}
\usepackage{textcomp}
\usepackage{xcolor}
\usepackage{siunitx}
\usepackage[english]{babel}
\hypersetup{hidelinks}

\title{GPT-5 vs Other LLMs in Long Short-Context Performance\thanks{This is the preprint version of a paper accepted for publication in the 3rd International Conference on Foundation and Large Language Models (FLLM2025). © IEEE. The final version will be available in IEEE Xplore.}}

\date{}
\author{ \href{https://orcid.org/0000-0003-4773-4370}{\includegraphics[scale=0.06]{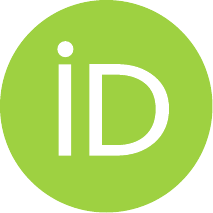}\hspace{1mm}Nima Esmi}\\
	Bernoulli Institute\\
	RUG, Groningen, The Netherlands\\
	ISRC, Khazar University, Baku, Azerbaijan\\
	\url{n.esmi.rudbardeh@rug.nl} \\
	\And
	\href{https://orcid.org/0009-0002-1323-0908}{\includegraphics[scale=0.06]{orcid.pdf}\hspace{1mm}Maryam Nezhad-Moghaddam}\\
	Department of Computer Engineering\\
	University of Guilan\\
	Rasht, Iran\\
	\url{maryam.n.moghaddam@gmail.com} \\
	\And
	\href{https://orcid.org/0009-0002-4823-276X}{\includegraphics[scale=0.06]{orcid.pdf}\hspace{1mm}Fatemeh Borhani}\\
	Department of Computer Engineering\\
	University of Guilan\\
	Rasht, Iran\\
	\url{f.borhani.m@gmail.com} \\
	\AND
	\href{https://orcid.org/0000-0002-5195-1688}{\includegraphics[scale=0.06]{orcid.pdf}\hspace{1mm}Asadollah Shahbahrami}\\
	Department of Computer Engineering\\
	University of Guilan, Rasht, Iran\\
	ISRC, Khazar University, Baku, Azerbaijan\\
	\url{shahbahrami@guilan.ac.ir} \\
	\And
	\href{https://orcid.org/0009-0005-9386-1321}{\includegraphics[scale=0.06]{orcid.pdf}\hspace{1mm}Amin Daemdoost}\\
	Department of Computer Engineering\\
	University of Guilan\\
	Rasht, Iran\\
	\url{amindaemdoost@gmail.com} \\
	\And
	\href{https://orcid.org/0000-0002-3678-7007}{\includegraphics[scale=0.06]{orcid.pdf}\hspace{1mm}Georgi Gaydadjiev}\\
	QCE Department\\
	TU Delft\\
	Delft, The Netherlands\\
	\url{g.n.gaydadjiev@tudelft.nl}
}

\renewcommand{\headeright}{This is the preprint version of a paper accepted for publication in the FLLM2025.}
\renewcommand{\shorttitle}{}

\begin{document}
\maketitle

\begin{abstract}
	With the significant expansion of the context window in Large Language Models (LLMs), these models are theoretically capable of processing millions of tokens in a single pass. However, research indicates a significant gap between this theoretical capacity and the practical ability of models to robustly utilize information within long contexts, especially in tasks that require a comprehensive understanding of numerous details. This paper evaluates the performance of four state-of-the-art models (Grok-4, GPT-4, Gemini 2.5, and GPT-5) on long short-context tasks. For this purpose, three datasets were used: two supplementary datasets for retrieving culinary recipes and math problems, and a primary dataset of 20K social media posts for depression detection. The results show that as the input volume on the social media dataset exceeds 5K posts (70K tokens), the performance of all models degrades significantly, with accuracy dropping to around 50-53\% for 20K posts. Notably, in the GPT-5 model, despite the sharp decline in accuracy, its precision remained high at approximately 95\%, a feature that could be highly effective for sensitive applications like depression detection. This research also indicates that the ``lost in the middle'' problem has been largely resolved in newer models. This study emphasizes the gap between the theoretical capacity and the actual performance of models on complex, high-volume data tasks  and highlights the importance of metrics beyond simple accuracy for practical applications.
\end{abstract}


\section{Introduction}

Large Language Models (LLMs) have seen a dramatic expansion in their context window capabilities, with recent models theoretically capable of processing millions of tokens in a single pass. This advancement promises to unlock new applications in long-document comprehension, enabling complex analysis of legal texts, financial reports, extensive codebases, and health records. However, a growing body of research reveals a significant disconnect between the theoretical context length and the practical ability of these models to robustly utilize the information contained within it \cite{yang2024large,han2024xbrl,10.1145/3731753,guevara2024large,esmi2025stress}.

\subsection{Prior Literature}
\begin{figure}
\centerline{\includegraphics[width=1\linewidth]{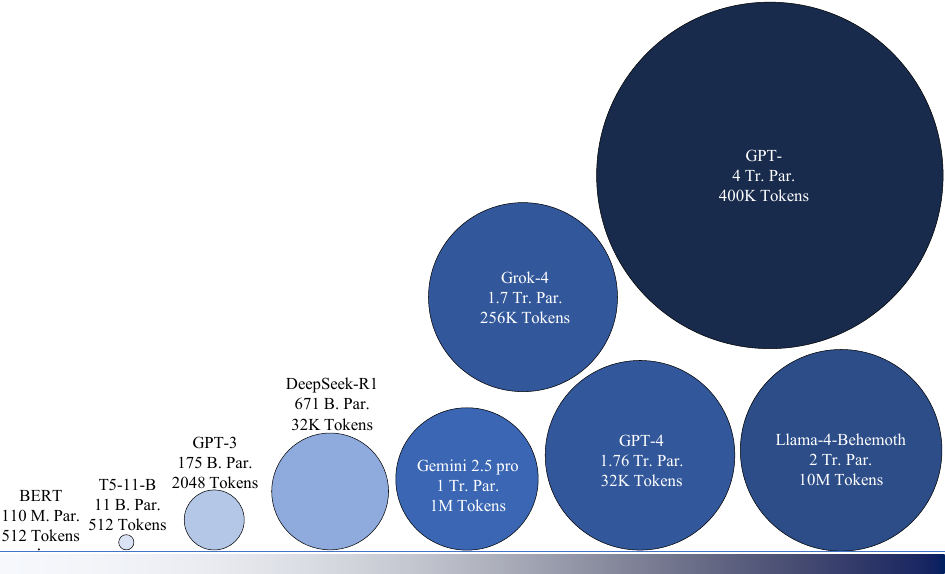}}
\caption{Growth in the number of parameters of large language models surpassing the trillion mark, alongside the expansion of their context windows..}
\label{ullms}
\end{figure}

As shown in Figure \ref{ullms}, with the number of parameters in LLMs surpassing the trillion mark, it is claimed that these models are capable of comprehending long texts, such as multi-hundred-page books, in a single pass.
%%%%%%%%%%%%%%%%%%%%% Figure %%%%%%%%%%%%%%%%%%%%%
%%%%%%%%%%%%%%%%%%%%% Figure %%%%%%%%%%%%%%%%%%%%%
%%%%%%%%%%%%%%%%%%%%% Figure %%%%%%%%%%%%%%%%%%%%%
%%%%%%%%%%%%%%%%%%%%% Figure %%%%%%%%%%%%%%%%%%%%%
Recent studies have demonstrated that LLM performance can degrade substantially when critical information is located in the middle of a long context, a phenomenon often described as the ``lost in the middle'' problem. This issue highlights how models exhibit primacy and recency biases, performing best when relevant information is at the beginning or end of the input context, while struggling to access and utilize data in the middle, as evidenced in tasks like multi-document question answering and key-value retrieval~\cite{b1}. While standard benchmarks like ``needle in a haystack'' have been instrumental in identifying this issue, they primarily test for simple fact retrieval and may not fully capture the challenges of tasks requiring a holistic understanding of a multitude of granular details, such as a document containing a large number of social media posts~\cite{wu2024loongserve}.
Further research has explored these limitations through novel benchmarks and analytical frameworks. For instance, evaluations using literary texts, such as novels, reveal that LLMs fail to maintain stable understanding beyond 64k tokens, particularly in tasks involving plot summarization, story-world configuration, and narrative time estimation, suggesting that current benchmarks like ``lost in the middle'' do not fully assess complex long-context integration~\cite{hamilton2025too}. Similarly, studies on extreme-label classification tasks with input lengths up to 50k tokens and label spaces of 28 to 174 classes show that long-context LLMs struggle with in-context learning, exhibiting biases toward later-presented labels and poor reasoning over distributed information, as tested in benchmarks like LongICLBench~\cite{li2024long}.
Additional investigations emphasize that a large context window does not guarantee flawless analysis of long sequences. Empirical analyses across models like Claude 3, GPT-3.5 Turbo, Gemini, Llama 3, and Mistral demonstrate performance degradation on tasks such as sentiment analysis and news categorization for lengthy inputs, with proposed solutions like extractive summarization pipelines improving accuracy by up to 50\% while reducing latency and costs~\cite{hosseini2024efficient}. From a mechanistic perspective, decomposing positional vectors in hidden states reveals how out-of-distribution positional encodings beyond the context window cause degradation, with methods like positional vector replacement and attention window extension enabling training-free length extrapolation~\cite{dong2024exploring}.

\subsection{Contribution}
This work underscores a critical gap in understanding how LLMs handle long sequences in realistic, fine-grained tasks, motivating our investigation into performance thresholds and model comparisons on specialized datasets. We aim to move beyond simple retrieval and understand the specific conditions under which model performance falters. Our research is guided by two central questions:
\begin{enumerate}
    \item \textbf{RQ1:} \textit{ How does increasing the length of long short-context inputs affect the analytical performance of Large Language Models in tasks involving fragmented and heterogeneous information?} We investigate the extent to which models preserve performance accuracy as the amount of fragmented and irrelevant information increases, potentially revealing degradation thresholds in holistic comprehension tasks.

    \item \textbf{RQ2:} \textit{ Which language models demonstrate higher precision in a specific application?} We will quantitatively compare the precision of selected LLMs (more specifically, Grok-4, GPT-4, Gemini, and GPT-5) on a clearly defined task and examine whether differences in architecture, context-window size produce systematic effects on measured precision.

\end{enumerate}

To empirically evaluate these questions, we have used three datasets designed to probe specific long-context failure modes. The primary dataset is a mental-health corpus of social-media posts labeled into two classes — depressed and not depressed — and is intended to evaluate classification and detection performance in a realistic, noisy domain. In addition, we created two supplementary, self-generated datasets that isolate retrieval and identification behaviors over long contexts: (1) a culinary recipes dataset of 1K recipes labeled vegetarian vs non-vegetarian (20K tokens total), where the model’s task is to identify all vegetarian options; and (2) a mathematical-problems dataset of 1K problems spanning five distinct categories (20K tokens total), where the model must retrieve all problems belonging to a specified category. This provide a robust framework for quantifying limitations of current LLMs on long-context retrieval and classification and for guiding future evaluations of extended-context capabilities.

\section{Datasets and Prompts}

A summary of the features of the datasets used and the objectives of their use is presented in Table \ref{tab:datasets}

%%%%%%%%%%%%%%%%% Table %%%%%%%%%%%%%%%%%
%%%%%%%%%%%%%%%%% Table %%%%%%%%%%%%%%%%%
\sisetup{
    table-number-alignment = center,
    table-format=5.0
}

\begin{table}[!t]
\centering
\caption{Summary of Datasets Used in This Study}
\label{tab:datasets}
\begin{tabular}{l c c c l}
\toprule
\textbf{Dataset} & \textbf{\#Samp.} & \textbf{Tokens} & \textbf{Class.} & \textbf{Goal} \\
\midrule
Depress.-Twitter & 20K & 300K & 2 & Depression detection \\
Culinary Recipes    & 1K  & 20K  & 2 & Vegetarian retrieval \\
Math--Problems      & 1K  & 20K  & 5 & Prob. \& Stat. retrieval \\
\bottomrule
\end{tabular}
\end{table}

%%%%%%%%%%%%%%%%% Table %%%%%%%%%%%%%%%%%
%%%%%%%%%%%%%%%%% Table %%%%%%%%%%%%%%%%%

We employed three datasets to investigate long-context performance degradation in LLMs, consisting of two supplementary datasets and one primary dataset.

\subsection{Supplementary Dataset 1 — Culinary Recipes}
This self-generated dataset contains 1K recipes ($\sim$20K tokens), including 100 vegetarian and 900 non-vegetarian entries. The classification task aims to identify all vegetarian recipes. The prompt provided to the LLMs was:

\textit{
In the uploaded text, among ``vegetarian'' and ``non-vegetarian'' find and sort the numbers of vegetarian recipes. Output only the vegetarian numbers, without any explanation.\\
``Vegetarian recipes'' means: Dishes crafted from plant-based ingredients, offering diverse, nutrient-rich meals without meat.\\
``Non-vegetarian\,recipes''\,means:\,Dishes\,featuring\,meat,\,poultry, or seafood, delivering bold flavors and protein-packed meals.}

\subsection{Supplementary Dataset 2 — Math--Problems}
This self-generated dataset contains 1K math problems ($\sim$20K tokens) spanning five categories: 58 \textit{Discrete \& Logical Mathematics}, 172 \textit{Algebra \& Algebraic Structures}, 114 \textit{Analysis \& Calculus}, 91 \textit{Geometry \& Spatial Mathematics}, and 65 \textit{Probability \& Statistics}. The objective is to identify all problems in the \textit{Probability \& Statistics} category. The prompt was:

\textit{In the uploaded text, among ``Discrete \& Logical Mathematics'', ``Algebra \& Algebraic Structures'', ``Analysis \& Calculus'', ``Geometry \& Spatial Mathematics'', ``Probability \& Statistics'' find and sort the numbers of Probability \& Statistics problems. Output only the problem numbers, without any explanation.\\
``Discrete \& Logical Mathematics'' means: Set Theory, Logic, Discrete, Combinatorics, Number Theory.\\
``Algebra \& Algebraic Structures'' means: Algebra, matrices, vectors, arithmetic.\\
``Analysis \& Calculus'' means: Calculus, sequences, trigonometry, optimization.\\
``Geometry \& Spatial Mathematics'' means: Geometry.\\
``Probability \& Statistics'' means: Probability, statistics.}

\subsection{Primary Dataset — Depress.-Twitter}
The primary dataset comprises 20K Twitter posts, evenly split into two classes: \textit{depressed} (10K posts) and \textit{non-depressed} (10K posts). The corpus contains approximately 300K tokens in total. The goal is binary classification for depression detection \cite{kaggle_mentalhealth}. The LLMs were given the following prompt:

\textit{In the uploaded text, among “Discrete \& Logical Mathematics”, “Algebra \& Algebraic Structures”, “Analysis \& Calculus”, “Geometry \& Spatial Math- ematics”, “Probability \& Statistics” find and sort all the numbers of Probability \& Statistics problems. Output only the problem numbers, without any explanation.}

\section{Evaluation Results}
\subsection{Evaluated Models}

Table \ref{tab:models} summarizes the LLMs used in this paper. We evaluated the performance of four state-of-the-art LLMs with extended context windows: Grok-4, GPT-4, Gemini 2.5, and GPT-5. These models differ in terms of architecture, training data, number of parameters, and maximum context length.

%%%%%%%%%%%%%%%%%%% Table %%%%%%%%%%%%%%%%%%%
%%%%%%%%%%%%%%%%%%% Table %%%%%%%%%%%%%%%%%%%
\begin{table}[!t]
\sisetup{
    table-number-alignment = center,
    table-format=7.0
}
\centering
\caption{Summary of Evaluated LLMs}
\label{tab:models}
\begin{tabular}{l c c l}
\toprule
\textbf{Model} & \textbf{\#Parameters} & \textbf{Context Tokens} & \textbf{Developer} \\
\midrule
Grok-4\cite{x.ai_grok4}         & 1.7T & 256K   & xAI \\
GPT-4\cite{neoteric_llama3}          & 1.76T & 32K & OpenAI \\
Gemini 2.5\cite{google_gemini_pro}    & 1T & 1M & Google DeepMind \\
GPT-5\cite{openai_gpt5}            & 4T & 40K & OpenAI \\
\bottomrule
\end{tabular}
\end{table}

%%%%%%%%%%%%%%%%%%% Table %%%%%%%%%%%%%%%%%%%
%%%%%%%%%%%%%%%%%%% Table %%%%%%%%%%%%%%%%%%%

\subsubsection*{Grok-4}
Developed with 1.7 trillion parameters and supporting a maximum context length of 256K tokens, Grok-4 is designed for high-throughput processing and general-purpose reasoning across various domains \cite{x.ai_grok4}.
\subsubsection*{GPT-4}
Released by OpenAI with 1.76 trillion parameters and a 32,768-token context window, GPT-4 has demonstrated strong performance on reasoning and text understanding benchmarks, though it is limited in handling ultra-long contexts \cite{neoteric_llama3}.
\subsubsection*{Gemini 2.5}

Gemini 2.5 supports a 1 million token context window, enabling the processing of very large documents in a single pass. It is optimized for reasoning across heterogeneous and long-form data \cite{google_gemini_pro}.
\subsubsection*{GPT-5}
GPT-5 offers a 400K token context window and enhanced retrieval accuracy, particularly in scenarios involving long sequences with embedded distractor information. It represents a significant step forward in balancing reasoning ability and long-context retention \cite{openai_gpt5}.

The models were executed using the API's default parameter settings for \texttt{temperature}, \texttt{top-p}, and \texttt{frequency-penalty}, with the maximum context window supported by each model.

\subsection{Implementation Procedure}
As shown in Figure \ref{implementation}, the experimental workflow consisted of the following steps:

\begin{figure}
\centerline{\includegraphics[width=1\linewidth]{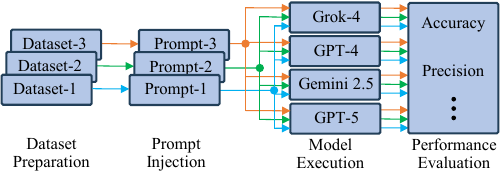}}
\caption{The workflow for our experiment, detailing the four primary phases: Dataset Preparation, Prompt Injection, Model Execution, and Performance Evaluation.}
\label{implementation}
\end{figure}

\subsubsection*{Dataset Preparation}
The three datasets described in Section~\ref{tab:datasets} were each divided into subsets to evaluate model performance with varying input lengths while maintaining original class distributions. The Culinary Recipes and Math-Problems datasets were subdivided into subsets of 125, 250, 500, and 1K posts/problems. In contrast, the Depress.-Twitter dataset, which is the primary dataset for this study, was divided into larger subsets of 5K, 10K, 15K, and 20K posts. This approach allows model performance evaluation at different scales of data complexity and length.

\subsubsection*{Prompt Injection}
For each subset, the corresponding prompt was appended to the dataset content and provided as input to the target LLM. The prompts instructed the models to identify:
\begin{itemize}
    \item Posts containing depression-related content in the \textit{Depression--Twitter} dataset.
    \item Vegetarian recipes in the \textit{Culinary Recipes} dataset.
    \item \textit{Probability\&Statistics}\,issues\,in\,the\,\textit{Math--Problem} dataset.
\end{itemize}

\subsubsection*{Model Execution}
Each LLM processed the inputs independently. The outputs consisted solely of the identifiers (indices) of the detected relevant posts, recipes, or problems, without any additional explanation. All models were executed using the API's default parameter settings for \texttt{temperature}, \texttt{top-p}, and \texttt{frequency-penalty}, with the maximum context window supported by each model.

\subsubsection*{Performance Evaluation}
The predictions generated by each model are compared against the ground-truth labels of the datasets. Standard classification metrics such as accuracy and precision are computed for each subset size and dataset. Precision is of utmost importance in certain applications. For instance, in detecting mental health disorders from posts shared by users on social media, a high level of precision in a model leads to increased confidence in positive diagnoses.

\section{Results}

In this section, the results obtained from the evaluation of the models on three different datasets are examined. First, the performance is analyzed from the perspective of accuracy and precision. Then, the quality of the model's performance along the dataset is investigated.

\begin{figure}
\centerline{\includegraphics[width=1\linewidth]{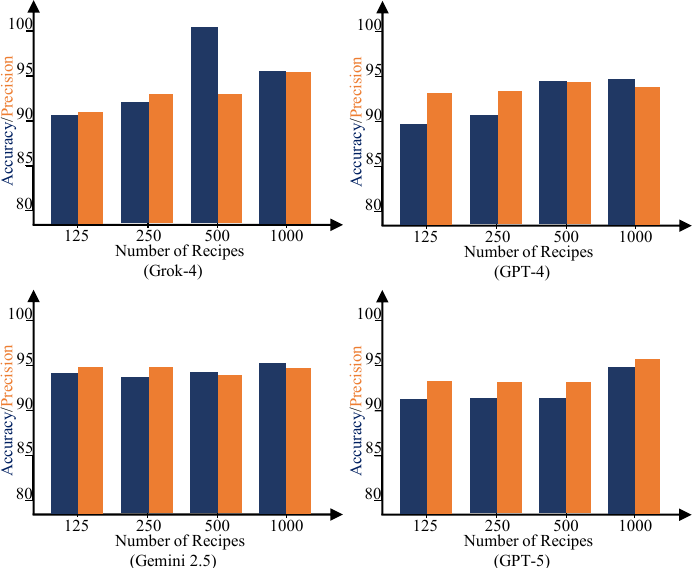}}
\caption{Evaluation results for the Culinary Recipes dataset across four subsets: 125, 250, 500, and 1K recipes.}
\label{recipe}
\end{figure}

Figure \ref{recipe} shows the average evaluation results of three repetitions of model performance on four subsets of 125, 250, 500, and 1K recipes from the perspective of accuracy and precision.
In many cases, as the number of recipes increases, not only do accuracy and precision not decrease, but a slight increase is also observed. Among them, GPT-5 had the best performance with an accuracy of more than 97\%.

\begin{figure}
\centerline{\includegraphics[width=1\linewidth]{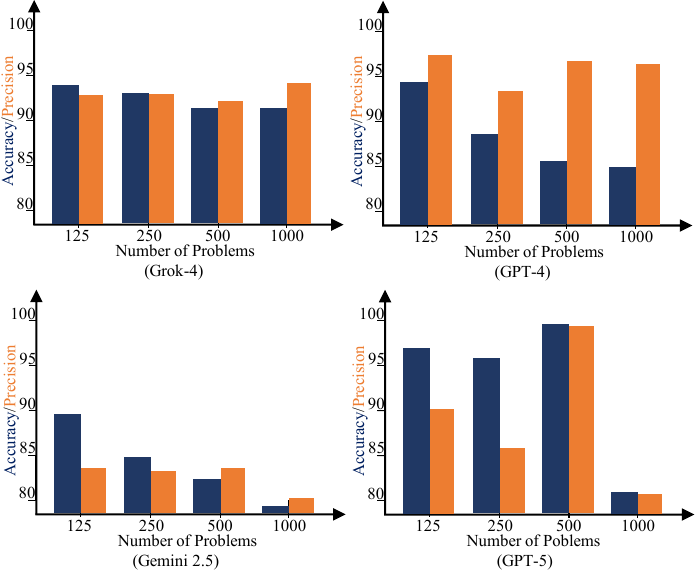}}
\caption{Evaluation results for the Math–Problem dataset across four subsets: 125, 250, 500, and 1K problems. The purple columns represent accuracy, while the orange columns represent precision.}
\label{math}
\end{figure}

Figure \ref{math} presents the evaluation of the models on the Math–Problems dataset. Unlike the recipes dataset, the models showed highly varied performance on the mathematical topics. Among them, Grok demonstrated consistent performance, while Gemini had a very weak performance. GPT-5 also showed significant fluctuations in its accuracy.

For the Depress.-Twitter dataset, the models behaved similarly to the Culinary Recipes dataset. However, for documents with more than five thousand tweets (more than 70K tokens), the models' performance observably decreases. This decline is significantly more pronounced in the Grok model. For higher volumes, up to 20K tweets (300K tokens), accuracy sharply decreases, reaching around 50 to 53 percent in binary classification. An interesting point, however, is that in GPT-5, despite the sharp decrease in accuracy, precision remains at around 95\%, which can be very effective for depression detection solutions.

\begin{figure}
\centerline{\includegraphics[width=1\linewidth]{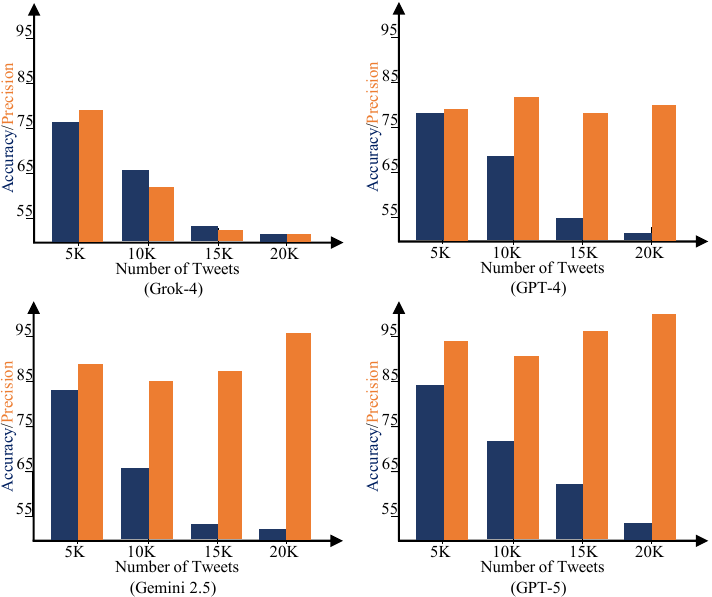}}
\caption{Evaluation results for the Depress.-Twitter dataset across four subsets: 5K, 10K, 15K, and 20K Tweets. The purple columns represent accuracy, while the orange columns represent precision.}
\label{depression}
\end{figure}

\begin{figure*}
\centerline{\includegraphics[width=1\linewidth]{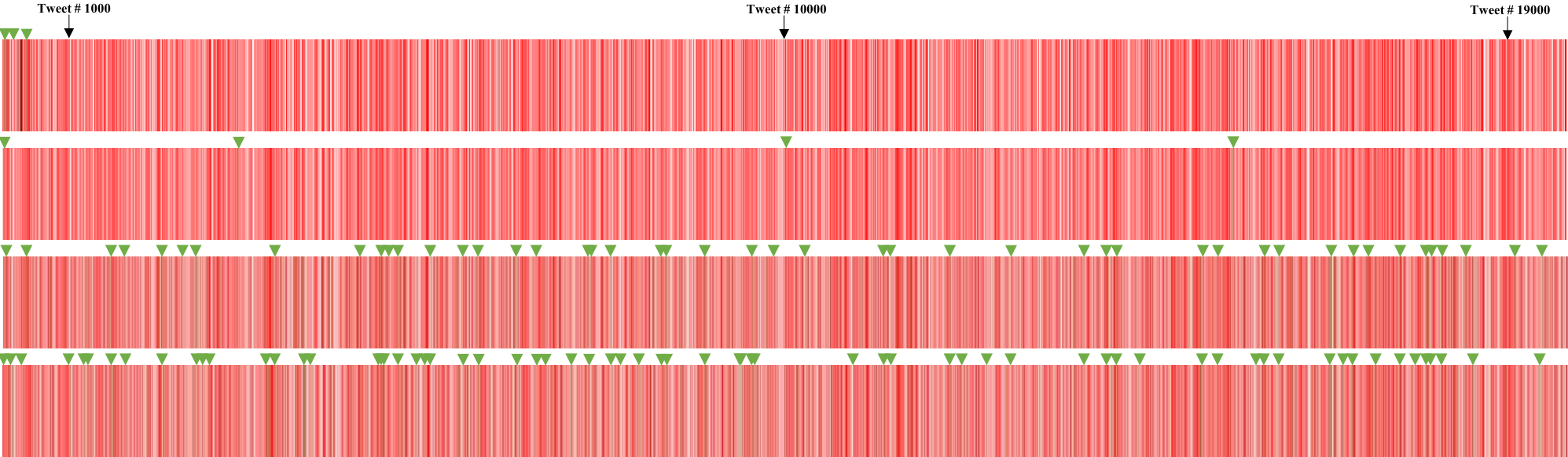}}
\caption{ The distribution of True Positives (green), True Negatives (white), False Positives (black), and False Negatives (red) for a single experiment on the entire Depress.-Twitter dataset. Green arrows serve as a guide to show the density of True Positives. The results from top to bottom belong to Grok, GPT-4, Gemini, and GPT-5, respectively. the purple columns represent accuracy, while the orange columns represent precision.}
\label{ullms-compare}
\end{figure*}

\begin{figure} 
\centerline{\includegraphics[width=1\linewidth]{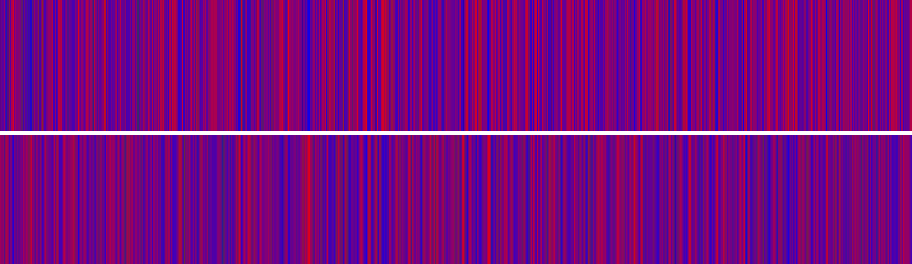}}
\caption{Differences in the performance of the Gemini model across two identical tests on a similar dataset, which yielded different results. The red lines indicate the density of false positives.}
\label{gemini-heatmap}
\end{figure}

Figure \ref{ullms-compare} shows the performance of Grok, GPT-4, Gemini, and GPT-5 from top to bottom. 
Here, white represents true negatives, green represents true positives, red represents false negatives, and black represents false positives. The green arrows are provided as a guide to indicate dense true positive spots. As can be seen from the figure, the "lost in the middle" problem is almost non-existent in the newer models. However, the Grok model found most of the depressive tweets at the beginning of the document.\\
Upon careful observation of the image and the location of the red and white markings, we can infer that there are relative similarities in the models' ability to detect or fail to detect depression in tweets. However, this was not consistent for Gemini across all repetitions of the same experiment.\\ For instance, consider Figure \ref{gemini-heatmap}. In this figure, areas with a high density of false negatives are shown in red. Below it, another density distribution is shown for false positives. As can be seen, the top and bottom red markings are complementary, suggesting that wherever there was a high false negative error in the first repetition, it was reduced in the second repetition, and vice versa.
This could be due to the initial settings of the attention mechanisms or changes in temperature, top-p, and frequency-penalty.

%%%%%%%%%%%%%%%%%%%%% Figure %%%%%%%%%%%%%%%%%%%%%
%%%%%%%%%%%%%%%%%%%%% Figure %%%%%%%%%%%%%%%%%%%%%
%

%%%%%%%%%%%%%%%%%%%%% Figure %%%%%%%%%%%%%%%%%%%%%
%%%%%%%%%%%%%%%%%%%%% Figure %%%%%%%%%%%%%%%%%%%%%

%%%%%%%%%%%%%%%%%%%%% Figure %%%%%%%%%%%%%%%%%%%%%
%%%%%%%%%%%%%%%%%%%%% Figure %%%%%%%%%%%%%%%%%%%%%

%%%%%%%%%%%%%%%%%%%%% Figure %%%%%%%%%%%%%%%%%%%%%
%%%%%%%%%%%%%%%%%%%%% Figure %%%%%%%%%%%%%%%%%%%%%

\section{Conclusion}

This study examined the performance of four state-of-the-art Large Language Models—Grok-4, GPT-4, Gemini 2.5, and GPT-5—on tasks involving extremely long short-contexts, revealing that despite advances in extending context window sizes, a substantial gap remains between models’ theoretical capacity and practical efficacy in complex, detail-rich applications. While modern LLMs can process hundreds of thousands of tokens, results showed a clear performance threshold beyond which comprehension deteriorates—specifically, inputs exceeding about 70K tokens in the depression detection dataset caused accuracy to drop to around 50–53\%. However, GPT-5 maintained exceptionally high precision (~95\%), suggesting that even when overall accuracy declines, some models retain reliability in sensitive domains such as mental health assessment. The findings also confirm that newer generations have largely mitigated the long-standing “lost in the middle” issue, reflecting progress in representing and retrieving contextual information across extensive sequences. These insights emphasize that evaluation should consider precision, robustness to fragmented information, and stability across input scales, while future LLM architectures should focus on enhancing information retention, contextual weighting, and multi-source coherence for more effective comprehension and reasoning over large-scale, heterogeneous data.

\end{document}